# Short-term Streamflow and Flood Forecasting based on Graph Convolutional Recurrent Neural Network and Residual Error Learning

Xiyu Pan[1]; Neda Mohammadi[1,2]; John E. Taylor[1, *]

[1] School of Civil and Environmental Engineering, Georgia Institute of Technology, 790 Atlantic Dr NW, Atlanta, GA, 30332, United States

[2] Faculty of Engineering, The University of Sydney, NSW 2006, Australia

[*] Corresponding author at: School of Civil and Environmental Engineering, Georgia Institute of Technology, 790 Atlantic Dr NW, Atlanta, GA, 30332, United States.

E-mail address: jet@gatech.edu (John E. Taylor).

Accurate short-term streamflow and flood forecasting are critical for mitigating river flood impacts, especially given the increasing climate variability. Machine learning-based streamflow forecasting relies on large streamflow datasets derived from rating curves. Uncertainties in rating curve modeling could introduce errors to the streamflow data and affect the forecasting accuracy. This study proposes a streamflow forecasting method that addresses these data errors, enhancing the accuracy of river flood forecasting and flood modeling, thereby reducing flood-related risk. A convolutional recurrent neural network is used to capture spatiotemporal patterns, coupled with residual error learning and forecasting. The neural network outperforms commonly used forecasting models over 1-6 hours of forecasting horizons, and the residual error learners can further correct the residual errors. This provides a more reliable tool for river flood forecasting and climate adaptation in this critical 1-6 hour time window for flood risk mitigation efforts.

# 1 Introduction

Global river floods and the associated risks are one of the major impacts of climate change (Winsemius et al. 2016). There has been extensive research revealing the exacerbation and changes in the flooding possibilities (Alifu et al. 2022), magnitude (Li et al. 2023), frequency (Tarasova et al. 2023), duration (Rohde 2023), and timing (Blöschl et al. 2017), caused by climate changes in the form of heat, land cover, precipitation, etc., in the western U.S. (Touma et al. 2022), East Asia (Li et al. 2023), Europe (Blöschl et al. 2017), Africa (Marshall 2023), and other parts of the world. The impact of climate change is projected to exacerbate human loss due to river floods by 70–83% and economic losses by 160–240% with a temperature increase of 1.5 °C (Dottori et al. 2018), and the regional direct losses caused by these floods can propagate within the global trade and supply network with more profound impacts (Willner et al. 2018). In response to increased river flood risks, adaptation strategies such as reducing flood peaks using detention areas, strengthening protection through dyke systems, river flood modeling, and population relocation have been studied (Dottori et al. 2023). Accurate short-term (i.e., hourly and daily (Wang et al. 2023)) streamflow forecasting is the input for many flood mitigation operations (Najafi et al. 2024), informing the storage and release of water in detention areas, dyke systems reinforcement ahead of high flows, flood simulation and risk anticipation, etc.

  A significant trend in streamflow forecasting is the prevalence and rapid growth of machine learning, especially deep learning approaches (Cui et al. 2022; Fang et al. 2023; Nguyen et al. 2022), which have demonstrated great potential in implicitly capturing complex temporal and spatial patterns (Fang et al. 2023; Habert et al. 2016; Niu and Feng 2021). Most machine learning-based streamflow forecasting studies employed supervised learning models and conducted the training in an end-to-end manner. This paradigm relies on large amounts of streamflow data as the ground truth for computing loss function values. Because streamflow is difficult to measure directly, in most relevant studies, the large streamflow ground truth dataset (hereafter referred to as reported streamflow) was derived from water level gauge heights using rating curves, which are fitted using water levels and a small number of field-measured streamflow (hereafter referred to as measured streamflow) data points (Hodson et al. 2024; Westerberg et al. 2011).

Despite their well-established modeling methods and widely supported effectiveness, rating curves are not immune to errors when estimating streamflow from measured water level data. As per the results of prior studies, the errors from rating curves are not negligible in many cases. In some studies, the error rates of rating curves were in the range of about 5% to 25% (Das Bhowmik et al. 2020), and some studies have reported error rates exceeding 50% (Bahreinimotlagh et al. 2019). Because machine learning-based short-term streamflow forecasting relies on large streamflow datasets as ground truth, errors introduced by rating curves in reported streamflow could affect the accuracy of streamflow forecasting. Therefore, even if a machine learning model for streamflow forecasting can make perfect forecasts of the reported streamflow, it still may fail to accurately inform corresponding flood management.

Errors in the reported streamflow due to rating curves can be more evident in flood streamflow forecasting scenarios. On the one hand, among the major reasons why rating curves may deviate from real-world conditions, anthropogenic (e.g., sand extraction, dam construction) and natural (e.g., weed growth) change of river channels can be addressed by updating rating curves every several weeks (U.S. Geological Survey 2024), and streamflow field measurement errors are often modeled as white noise (Coxon et al. 2015). However, errors from unsteady flows (e.g., abrupt streamflow change) and imperfect rating curve fitting (e.g., function form selection, extrapolation) (Coxon et al. 2015; Kiang et al. 2018; Wolfs and Willems 2014) are still prone to exist in rating curves. During flood rising, the water surface slope will be greater than the steady flow slope at the same water level, due to the abrupt and rapid increase in inflow. Thus, the streamflow would be greater than that indicated by the rating curve, and vice versa when a flood recession occurs (Wolfs and Willems 2014).

Although some prior studies attempted to reduce the error resulted from imperfect forecasting by machine learning models (Li et al. 2017, 2021), only a few studies have focused on reducing the error between the reported and measured streamflow (Das Bhowmik et al. 2020; Habert et al. 2016; Zeroual et al. 2016). In the study closest to the present one, Das Bhowmik et al.(Das Bhowmik et al. 2020) aimed to propose a method for reducing the error induced by rating curves for streamflow forecasting, but the error was modeled using hypothesized distributions and studied in a simulation environment. In this study, we aim to propose a method to reduce the impact of reported streamflow errors on machine learning-based hourly streamflow and flood forecasting. It first forecasts the reported streamflow and then further learns and reduces the errors regarding measured streamflow. The proposed method has two parts, where the first part is a streamflow forecasting base model. The base model is built on a graph convolutional recurrent neural network, and it uses the water level at gauging stations and the rainfall in the watershed at past time points to forecast the reported streamflow. For the base model, the forecasting error is computed as the difference between the forecasts and the reported streamflow, while in the second part of the method, the error is computed regarding the measured streamflow and referred to as residual error. The river water level and the change in water level were used to learn and forecast the residual error, which was then utilized to modify the initial streamflow forecasts. The proposed method is expected to reduce the impact of reported streamflow data errors on short-term streamflow forecasting, improve the accuracy of forecasting, especially during flood periods, and contribute to relevant flood management decisions (e.g., water releases) that rely on streamflow estimates.

## 2 Literature review

### *2.1 Short-term streamflow forecasting based on machine learning*

The proposed research builds on previous studies of machine learning-based short-term streamflow forecasting. In related studies, short-term forecasting usually refers to hourly streamflow forecasting one to six hours in advance (Kao et al. 2020; Lin et al. 2021; Nguyen et al. 2022; Zhang et al. 2022). Some studies have tried to extend the forecast horizon. In the work by Cui et al., (2022), the forecasting is made every three hours, with a three to 12-hour forecast horizon. Olfatmiri et al. (2022) tested the hourly forecasting performances under the 1, 2, 6, 12, 18, 24, and 48 hours horizon, but the forecasting accuracy decreases quickly with increasing horizons. On the other hand, very few studies reduced the forecasting horizon to less than an hour (Kurian et al. 2020).

The relatively short horizon of streamflow forecasting influences the choice of machine learning model features. Unlike long-term streamflow forecasting, which often uses forecasted precipitation, soil moisture, and climatic factors, etc. as features, the most commonly used features for short-term streamflow forecasting are the flow information (e.g., streamflow, water level) and rainfall at past time points (Cui et al. 2022; Kao et al. 2020). To simplify the inputs to the machine learning model, rainfall is aggregated to the average of the data from multiple upstream rainfall measurement stations in most prior studies (Lin et al. 2021). A small number of studies used other features, such as maximum and minimum 2m daily air temperatures (Zhang et al. 2022), and rainfall forecasts (Kurian et al. 2020). A notable feature is the streamflow or water level from the upstream gauging stations. In prior studies, many of the studied locations did not have upstream gauging stations, but where they do exist, including the upstream information as an input to the machine learning model is a favorable modeling practice for better accuracy (Nguyen et al. 2022).

Many machine learning models have been used to extract patterns from features and targets. Neural network models have gained popularity in recent studies. The basic multilayer perceptron is often used as the baseline model for model comparison (Kao et al. 2020; Zhang et al. 2022). The suggested model is usually based on recurrent neural (RNN) network architecture. Because short-term future streamflow is highly correlated with the streamflow and rainfall at past time points, RNNs are a natural choice, whose architecture is designed to extract temporal patterns between features and targets. As one of the RNN variants, the Long Short-term Memory (LSTM) model was widely adopted (Cui et al. 2022). Different modeling techniques, such as encoder-decoder structure (Kao et al. 2020), and time series differencing (Lin et al. 2021), have been used to enhance the performance of the basic LSTM. In addition, a few studies have applied Transformer to the problem of short-term streamflow forecasting, considering its potential advantage of memorizing more historical information (Xu et al. 2023). However, although previous studies have used time series of streamflow and rainfall from multiple gauging stations in the watershed as model inputs, models designed to capture spatial patterns have rarely been used. Therefore, except for following the usual modeling practices of previous studies (i.e., 1–6 hour forecasting horizons, the use of streamflow and rainfall as features, and the adoption of the RNN architecture), our study adopts a graph convolution technique, inspired by prior research in spatiotemporal forecasting of urban systems (Wang et al. 2024; Wu et al. 2023; Zanfei et al. 2022). The graph convolution aggregates information from gauging stations in the watershed, enabling the model to capture spatial dependencies crucial for a more realistic representation of the hydrological system and more accurate streamflow forecasting at the target location.

*2.2 Streamflow forecasting with residual error learning*

Previous studies have considered the residual error in streamflow forecasting. The residual error is defined as the difference between the forecasted streamflow and the streamflow working as the ground truth. It comes from the uncertainties in hydro-climatological data measurements, the errors in the forecasting model inputs, the flawed parameters of the forecasting model, the imperfect design of the conceptual model, etc. (Panchanathan et al. 2023). Not only in hydrologic models, but these sources of residual error may also exist in precipitation forecasting models for streamflow forecasting. For medium- and long-term streamflow forecasts, the residual error from the precipitation forecasting models is also a major concern in previous studies (Khatun et al. 2023; McInerney et al. 2020).

A foundational assumption by the studies on residual error learning and reduction is that the residual errors could be described by explicit or sometimes implicit patterns, such as biases, variances, autocorrelations, and distributions. While various attempts have been made to identify and decompose the errors into explainable sources, many studies in recent years primarily consider the overall forecasting errors and their underlying patterns (Li et al. 2016). For instance, McInerney et al. (2020) managed to model and forecast the residual error considering seasonal variability, dynamic biases, and non-Gaussian residual errors. Roy et al. (2023) assumed that residual error is not a white noise following Gaussian distribution, and it is time-dependent or has other definite patterns.

While the previous study managed to forecast both streamflow and residual errors with a single model by redesigning the structure of regression models considering the residual error patterns (McInerney et al. 2020), most relevant studies relied on an additional residual error module beyond the base model. For example, Khatun et al. (2023) used LSTM to model and forecast the residual error, and Roy et al. (2023) attached a random forest model to learn the residual error beyond the base simulation model. M. Li et al. (2016, 2017) proposed an alternative strategy of residual error learning. Instead of using a complex or deep model to learn the residual error, applying multiple simple regression models at consecutive stages is possible to obtain better results. After applying a sine and logarithmic transformations on the residual errors, they used the average of forecasted streamflow, the residual error in the last time point, and a mixture Gaussian distribution to model and reduce the error.

Previous studies of streamflow forecasting with residual error learning have established a deep understanding of residual errors and ways to deal with them. However, the residual errors regarding measured streamflow are rarely studied. A few studies have focused on this problem (Das Bhowmik et al. 2020; Habert et al. 2016; Zeroual et al. 2016). Das Bhowmik et al. (2020) proposed a method for reducing the residual error regarding measured streamflow. They designed some scenarios of errors and the corresponding error reduction and investigated the benefits of the "knowledge of error" in forecasting streamflow in a simulated environment, instead of using a real-world dataset. With the same research objective, Habert et al. (2016) modeled the rating curve error due to temporal change of friction coefficients and bed geometry in the river channel, using data assimilation. However, they did not consider other sources of rating curve error, such as unsteady flow and imperfect fitting.

## 3 Methods

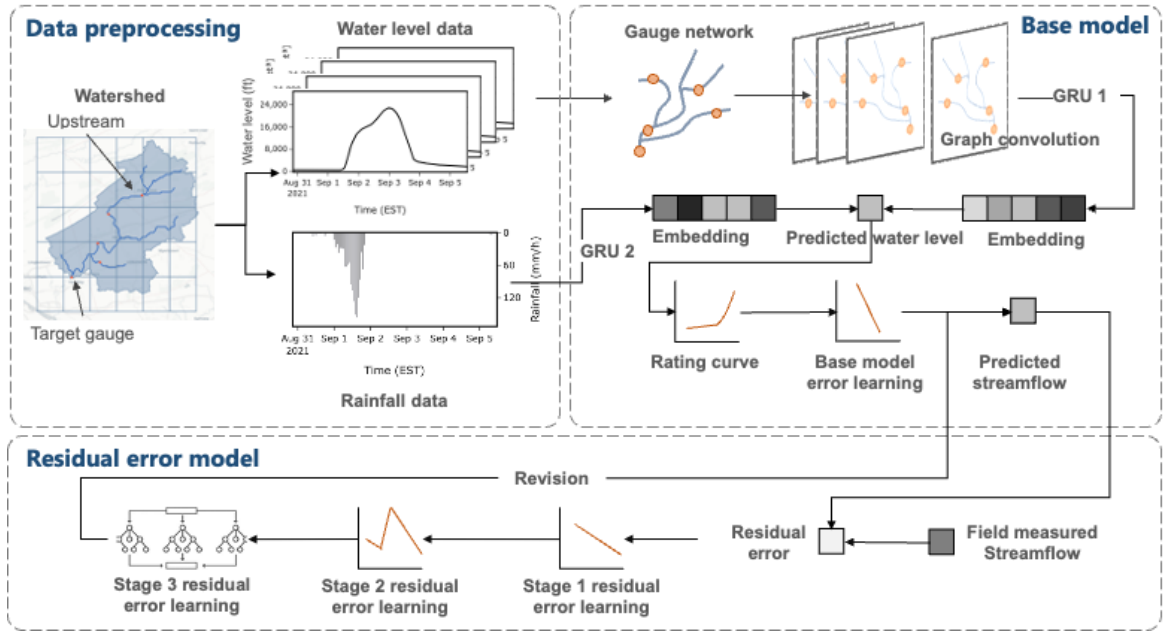

**Fig. 1.** Analytical framework of the present study. The modeling and analysis begin with data preprocessing, which supports the modeling of the Gated Recurrent Unit (GRU) and the graph convolution in the base model. The output of the base model is revised by three stages of error learning, before being used for streamflow and flood forecasting.

In response to the shortcomings of prior studies, this study proposes a method based on machine learning models for short-term streamflow and flood forecasting, with residual error learning for reducing the errors introduced by rating curves. As shown in Fig. 1, the proposed method follows the paradigm of previous studies, in which a base model is used together with an additional residual error learning module. In the base model, average rainfall in the watershed and the water level at the target gauging station as well as upstream stations at past time points are used as features, and the streamflow 1-6 hours later is the target. The base model is built on the Gated Recurrent Unit (GRU) structure (a variant of RNN) for capturing temporal patterns and the graph convolution is utilized to extract spatial patterns in the data. The residual error learning part adopts the strategy suggested by M. Li et al. (2016). Instead of using a single complex model, multiple linear regression models were used to learn the errors introduced by unsteady flows and the imperfect fitting of rating curves, respectively. Inspired by the work of Roy et al. (2023), an XGBoost model is attached to the end of the process to learn the non-linear residual error patterns that are not covered by the linear models. The three stages of residual error learning were implemented sequentially. The model in the following stage is built on the data processed by the last stage model. The learned and forecasted residual error is used to revise the original forecasting by the base model and improve the forecasting.

*3.1 Data sources and preprocessing*

To support the basic idea of this study and to validate the effectiveness of the proposed method, multiple public datasets were collected and analyzed. We first implemented a large-scale data analysis to measure the magnitude of differences between reported and measured streamflow in the U.S. streamflow databases. The USGS operates one of the largest stream gauge networks within the U.S. and its data are widely used in machine learning-based streamflow prediction studies. Therefore, we started the analysis with all 10,754 stream gauges operated by the USGS that were active on Dec. 31, 2023 (waterdata.usgs.gov/nwis).

Because the motivation of this study is focused on forecasting streamflow and its impacts during floods and high flows, 3,488 surface water gauges monitoring medium-to-large streams were considered, and 1,548 out of them that experienced at least one flood condition during recent years were selected. Specifically, 1) streams with long-term streamflow greater than $10 m^3/s$ (i.e., streams rated higher than Level 5) in the HydroSHEDS V1 dataset (www.hydrosheds.org) were identified as medium-to-large streams, and the corresponding USGS gauges on these streams were identified based through spatial join. 2) To determine whether streams are prone to flooding and high flows in recent years, instantaneous (i.e., 15 minutes time interval) gauge height values from Jan. 1, 2018, to Dec. 31, 2023, were retrieved from the USGS NWIS and resampled to obtain the maximum daily gauge height values for the selected gauges. 3) Thresholds of flood stages determined by NOAA were downloaded from USGS WaterWatch (waterwatch.usgs.gov). Among the five flood stages (www.weather.gov/lot/hydrology_definitions), the "action stage" was selected as the threshold due to its action and decision-making significance.

For the selected gauging stations, streamflow field measurements and instantaneous reported streamflow between January 1, 2018, and December 31, 2023, were collected from the NWIS. 1) Field measurements are streamflow values measured on-site by USGS hydrologic technicians every few weeks or months using equipment such as the current meter or Acoustic Doppler Current Profiler. Missing values in the raw data and values that were not used in rating curve development were excluded from the analysis. 2) Instantaneous reported streamflow was modeled by the USGS using gauging heights and the temporary rating curve used at the corresponding period. Provisional values were removed from the reported streamflow. The pre-processed reported and measured streamflow data were used to calculate the differences between the two and it is analyzed in the Results section.

To implement and validate the proposed method, candidate gauging stations were further filtered based on criteria such as multiple flooding events in recent years, multiple stations within the watershed, a minimum number of streamflow field measurements at the target station during the test period, and sufficient water level and streamflow data. The proposed method was tested on two gauging stations, which matches the number of test sites, i.e., one to two, in most prior relevant studies (Cui et al. 2022; Lin et al. 2021; Nguyen et al. 2022). USGS 01573560 is located at Swatara Creek near Hershey, Pennsylvania, U.S.A., which is one of the upstream rivers of a medium-sized city, Harrisburg, PA. The annual average streamflow in 2023 is 572 cubic feet per second and the drainage area is 483 square miles. USGS 03320000 is located at Green River at Calhoun, Kentucky, U.S.A. It is on the major upstream river of Evansville, Indiana, U.S.A. The average streamflow of the gauge's location in 2021 is 12,300 cubic feet per second, and the drainage area is 7,566 square miles.

For the selected gauging stations, 1) instantaneous water levels, instantaneous reported streamflow, and measured streamflow from Jan. 1, 2004, through Dec. 31, 2023, were collected from the NWIS and were either down-sampled or rounded to hourly data. 2) The geographic information required includes river channels, gauging stations, and watersheds. The geographic data for river channels is from HydroSHEDS V1. The geographic locations of gauging stations, as well as upstream gauging stations, were collected from the USGS Hydro Network-Linked Data Index (waterdata.usgs.gov/blog/nldi-intro). Geographic information of watersheds was collected from the USGS StreamStates database (streamstats.usgs.gov/ss/). 3) In addition, rainfall data were downloaded from the JAXA Global Rainfall Watch (GSMaP) database (sharaku.eorc.jaxa.jp/GSMaP/). The rainfall data has the same time frame as the streamflow data and is geographically scoped within the bounding box of the watershed.

*3.2 Graph convolution recurrent neural network for streamflow forecasting*

This section describes the base model for streamflow forecasting. In the base model, average rainfall in the watershed and the water level at the target gauging station as well as upstream stations at past time points are used as features, and the streamflow 1-6 hours later is the target. The base model is built on the Gated Recurrent Unit (GRU) structure (a variant of RNN) for capturing temporal patterns and the graph convolution is utilized to extract spatial patterns in the data. The predicted water level is converted to the predicted streamflow using the rating curves from the USGS for the corresponding periods.

To make a formal problem formulation, the studied watershed is conceptualized as a weighted and directed graph $\mathcal{G} = (\mathcal{V}, \mathcal{E}, \boldsymbol{A})$, where $\mathcal{V}$ ($|\mathcal{V}| = N$) is the set of nodes representing gauging stations, $\mathcal{E}$ is the connectedness between nodes, and $\boldsymbol{A} \in \mathbb{R}^{N \times N}$ is the weighted adjacency matrix of nodes. Each entry $A_{i,j}$ represents a node proximity, which is a function of the traveling distance in the river network between each pair of connected nodes. It is calculated as $A_{i,j} = \frac{1}{e^{distance(i,j)}}$ (Jia et al. 2022), after standardizing the traveling distances using the min-max method. The features from gauging stations at one time point are represented by $\boldsymbol{X_W} \in \mathbb{R}^{N \times D_{in,W}}$, where $D_{in,W}$ is the corresponding number of input features. In this study, $D_{in,W}$ is equal to one (i.e., water level). The total rainfall at one time point is denoted as $\boldsymbol{X_R} \in \mathbb{R}^{1 \times D_{in,R}}$. In this study, $D_{in,R}$ is equal to one (i.e., rainfall). Let $\boldsymbol{X^t}$ represent the feature matrix at a specific time point $t$. The formulated problem to solve is to learn a function $\mathcal{F}_\Phi$ parameterized by the parameter set $\Phi$ that maps $T$ historical observations to the water level observation at the target gauging station $h$ hours in the future, given the graph $\mathcal{G}$ and input data:

$$[\boldsymbol{X_W^{t-T+1}}, \boldsymbol{X_R^{t-T+1}}, \cdots, \boldsymbol{X_W^t}, \boldsymbol{X_R^t}\,;\, \mathcal{G}] \xrightarrow{\mathcal{F}_\Phi} x_{W^*}^{t+h} \quad (1)$$

, where $x_{W^*}^{t+h}$ is the water level of the target node at time $t + h$. Note $x_{W^*}^{t+h}$ is then converted to the streamflow $x_{S^*}^{t+h}$ at the same location and same time point using USGS's rating curves after the forecasting. Note that the USGS updates the rating curves, so different rating curves were used in the conversion for the corresponding applicable periods.

The function $\mathcal{F}_\Phi$ (i.e., the base model) to learn has the structure to 1) encode the water level time series at the target and upstream nodes using a graph convolutional GRU, 2) encode the total rainfall within the watershed using a GRU, and 3) output the forecasted water level using a multilayer perceptron after concatenating the embeddings from the above two steps. The graph convolution process is based on the work by Y. Li et al.(Li et al. 2018). It could be first understood as an update to the feature matrices (e.g., $X_W$) before sending it to the computations of GRU. The mechanism of the graph convolution is characterized by a random walker traveling between the nodes in the graph $\mathcal{G}$ to collect information from neighborhood nodes for the target node, and the random walker follows the probabilities of choosing the node in the next layer defined by the transition matrices. The example of the convolution on $X_W$ is as below.

$$X_{W; :, d_{in}} \star_{\mathcal{G}} \theta = \sum_{k=0}^{K-1} \theta_k (D_I^{-1} A^\top)^k X_{W; :, d_{in}} \quad \text{for } d_{in} \in \{1, \cdots, D_{in}\} \tag{2}$$

, where $\star_{\mathcal{G}}$ is the graph convolution operator, $K-1$ is the number of movements made by the random walker. The learnable parameters of the convolution are denoted as $\theta \in \mathbb{R}^K$. $\mathbf{1} \in \mathbb{R}^N$ is a vector of ones, and $D_I = diag(A^\top \mathbf{1})$ is the in-degree diagonal matrix. The diagonal values of $D_I$ indicate the number of upstream stations for each of the gauging stations. $(D_I^{-1} A^\top)^k$ represents the transition matrix of the random walker for each movement. Note that because water only flows from upstream to downstream in this study, and the status of a gauging station is only related to its upstream river, only in-degree is considered in the equation, and the out-degree matrix in Y. Li et al.(Li et al. 2018) work is omitted.

The convolution operation defined above does not change the dimension of the feature space, but a complete convolution layer should enable expanding the feature space dimension. Denote the parameter tensor for the convolution layer as $\Theta \in \mathbb{R}^{D_{out} \times D_{in} \times K}$. $D_{in}$ and $D_{out}$ are the dimensions of input and output feature space. The convolution at each movement of the random walker is parameterized by $\Theta_{d_{out}, d_{in}, :} \in \mathbb{R}^K$ for the $d_{in}^{-th}$ input feature and $d_{out}^{-th}$ output feature. The formal formulation is as below.

$$\boldsymbol{H}_{W;\,:,d_{out}} = a\left(\sum_{d_{in}=1}^{D_{in}} \boldsymbol{X}_{W;\,:,d_{in}} \star_{\mathcal{G}} \boldsymbol{\Theta}_{d_{out},d_{in},:}\right) \quad \text{for } d_{out} \in \{1,\cdots,D_{out}\} \tag{3}$$

, where $\boldsymbol{H}_{W;\,:,d_{out}}$ is the output feature matrix, and $\boldsymbol{a}$ is the activation function. $\boldsymbol{a}$ is set to the Rectified Linear Unit (ReLU), an activation function that outputs the input if it is positive and zero otherwise, in this study.

The graph convolution is used together with GRU to capture both the spatial and temporal patterns in the water level time series. In the basic GRU proposed by Chung et al.(Chung et al. 2014), the input $\boldsymbol{X}^t$ at time point $t$ and the hidden state $\boldsymbol{H}^{t-1}$ from the last time point are linearly transformed separately when calculating the reset gate, the update gate, and the candidate hidden state. In the graph convolutional GRU, the $\boldsymbol{X}^t$ and $\boldsymbol{H}^{t-1}$ are concatenated first, then convoluted by the graph convolution layer defined above. The rest of the computation of the basic GRU remains the same. Readers can refer to the work by Y. Li et al.(Li et al. 2018) for the complete formulation. The graph convolutional GRU defined here is used to process historical water level time series, and it generates the embedding of all inputs at the last time point. Denote the output embedding of the graph convolutional GRU as $\boldsymbol{H}_W^t$. Note that $\boldsymbol{H}_W^t$ contains the embedded information for all nodes, and it needs to be sliced for extracting the information belonging to the target gauging station.

The graph convolutional GRU mentioned above is redundant for processing total rainfall $\boldsymbol{X}_R$ within the watershed since $\boldsymbol{X}_R$ omits the spatial distribution of rainfall. Therefore, the basic GRU model was used to process the rainfall time series. Like the behavior of the graph convolutional GRU, the GRU would also produce the embedding of the rainfall at each time step. Denote the output embedding at the time $t$ as $\boldsymbol{H}_R^t$. $\boldsymbol{H}_R^t$ and $\boldsymbol{H}_W^t$ are concatenated after slicing $\boldsymbol{H}_W^t$ to obtain the information of the target node. A multilayer perceptron (MLP) with ReLU activations works as a decoder to transform the concatenated embedding to the forecasted water level $x_{W^*}^{t+h}$, as below.

$$x_{W^*}^{t+h} = MLP(\ concat(\ \boldsymbol{H_R^t},\ slice(\boldsymbol{H_w^t})\ )\ ) \tag{4}$$

The loss function for the model training is the weighted mean squared error calculated using $x_{W^*}^{t+h}$ and the ground truth water level, and the weight of a data point is determined by the density of its bin in the water level histogram and by a log transformation. After finishing the model training, the corresponding rating curve was used to convert $x_{W^*}^{t+h}$ to the forecasted streamflow $x_{S^*}^{t+h}$. Note that the conversion does not incur additional errors regarding the reported streamflow, as the reported streamflow published by USGS was generated by the same rating curves mentioned here.

To further investigate the accuracy of the base model, the errors of $x_{S^*}^{t+h}$ regarding the reported streamflow is calculated. Inspired by the work of M. Li et al.(Li et al. 2016), it is assumed that the error of the forecasting is autoregressive, and the error at time point $t$ is highly correlated with that at time point $t - h$. This characteristic is utilized to further reduce the error of the base model in forecasting the reported streamflow. By modifying the method by M. Li et al.(Li et al. 2016), a simple linear regression model for forecasting the base model error was designed, as shown below.

$$x_{S^*,reported}^t - x_{S^*}^t = \rho\ (x_{S^*,reported}^{t-h} - x_{S^*}^{t-h}) + b \tag{5}$$

, where $x_{S^*,reported}^t - x_{S^*}^t$ is the error at time $t$, and $x_{S^*,reported}^{t-h} - x_{S^*}^{t-h}$ is the error at time $t - h$. For the regression, it is assumed the training of this linear model could be implemented continuously and hourly in the real-world use case. Thus, for forecasting the error at $t + h$, all the forecasts by the base model and the reported streamflow at and before $t$ could be used to calculate the error and implement the regression. After the regression, the updated forecast of reported streamflow is $x_{S^*,updated}^{t+h} = x_{S^*}^{t+h} + \hat{\rho}(x_{S^*,reported}^t - x_{S^*}^t) + \hat{b}$.

For model training and testing, the period of the available data points (i.e., data points without null values) was split. The first 60% of the time was split for the training, the last 25% was split as the testing period, and the remaining 15% in the middle was used for the validation set. All the time series data was split according to the training, the validation, and the testing periods. During the model training phase, the batch size, the learning rate, the hidden state size, and the number of layers in the decoder MLP, were set as hyper-parameters and were determined by the Tree-structured Parzen Estimator (Watanabe 2023). The length of the time window $T$ was set to 24.

### *3.3 Residual error learning and reduction in stages*

The base model can forecast streamflow with a certain accuracy, and the ground truth for calculating the training loss is the reported streamflow. This is a reasonable practice, as the number of streamflow field measurements is very limited and is not sufficient to train deep learning, or even many shallow machine learning models. However, this practice incorporates the inaccuracies of rating curves into the model's training target. To address this issue, this section describes a method of utilizing the limited number of measured streamflow data for residual error learning and reduction and bridging the gap between the reported streamflow and the measured streamflow. The residual error learning part adopts the strategy suggested by M. Li et al.(Li et al. 2016). Instead of using a single complex model, multiple linear regression models were used to learn the errors introduced by unsteady flows and the imperfect fitting of rating curves, respectively. Inspired by the work of Roy et al.(Roy et al. 2023), an XGBoost model is attached to the end of the process to learn the non-linear residual error patterns that are not covered by the linear models. The three stages of residual error learning were implemented sequentially. The model in the following stage is built on the data processed by the last stage model. The learned and forecasted residual error is used to revise the original forecasting by the base model and improve the forecasting.

Prior to the implementation of Stage 1, the target of residual error learning was built by calculating the differences between the forecasted streamflow $x_{S^*,updated}^{t+h}$ and the measured streamflow $x_{S^*,measured}^{t+h}$. The differences are normalized using the forecasting itself to avoid residual errors at high water levels dominating the modeling.

$$r_1 = (x_{S^*,updated}^{t+h} - x_{S^*,measured}^{t+h}) / x_{S^*,updated}^{t+h} \tag{6}$$

, where $r_1$ is the error to learn in Stage 1. Among the sources of residual errors due to rating curves, Stage 1 considers the unsteady flow in the framework proposed by (Coxon et al. 2015). Specifically, Stage 1 aims to reduce the negative impact of hysteresis due to unsteady flows. Hysteresis is a phenomenon in which the response of a system to an external influence is affected "not only by the present magnitude of that influence, but also by the system history" (Kumar 2011). In the case of streamflow forecasting, hysteresis indicates that the relationship between streamflow and water levels is not only regarding the present water level but also regarding the water level at previous time points, when the rapid rise or fall in streamflow causes a change in the water surface gradient (Kumar 2011).

In the short-term streamflow forecasting, particularly considering floods, when streamflow increases or decreases rapidly, the effect of hysteresis is not negligible. Therefore, Stage 1 uses both the forecasted water level $x_{W^*}^{t+h}$ and the delta of the forecasted water level $\Delta x_{W^*}^{t+h} = x_{W^*}^{t+h} - x_{W^*}^{t+h-1}$ as features to forecast $r_1$. In contrast, the hysteresis effect is not significant when the flow is steady. To determine the thresholds for taking hysteresis effects into account, the flood stages determined by the NOAA for each of the gauging stations were utilized. Among the "action", "minor", "moderate", and "major" flood water levels, the "action" level $w_{action}$ is used for filtering. The measured streamflow with a water level below $w_{action}$ are filtered out. The ratio of the remaining number of measured streamflow over the total available measured streamflow is recorded as $e_{action}$, which is a percentage between 0% and 100%. Only the measured streamflow with a $x_{W^*}^{t+h}$ exceeding $w_{action}$ and with a $\Delta x_{W^*}^{t+h}$ in the top $e_{action}$ percent are kept for training a linear regression model $r_1 = \rho_1 \Delta x_{W^*}^{t+h} + b_1$. After the regression, $r_1$ is estimated as $\hat{r}_1 = \hat{\rho}_1 \Delta x_{W^*}^{t+h} + \hat{b}_1$.

Considering $r_1$ is not only influenced by the hysteresis, but also the error from the base model, a confidence index $c_1$ is designed to indicate the reliability of the $\hat{r}_1$ forecasted using the hysteresis rationale. The $x_{S^*,updated}^{t+h}$ at time points when field measurements are available are used to calculate the mean absolute percent error ($MAPE_{updated}$) with respect to the reported streamflow $x_{S^*,reported}^{t+h}$, and the mean absolute percent error (MAPE) of $x_{S^*,reported}^{t+h}$ regarding $x_{S^*,measured}^{t+h}$ at the same time points is denoted as $MAPE_{reported}$. $c_1$ is the calculated as the ratio of $MAPE_{updated}$ and $MAPE_{updated} + MAPE_{reported}$. The target of the Stage 2 residual error learning is therefore:

$$r_2 = r_1 - c_1 \hat{r}_1 \tag{7}$$

, where $r_2$ is the residual error after the improvement of Stage 1. And $x_{S^*,updated}^{t+h}$ is updated as

$$x_{S^*,1}^{t+h} = (1 - c_1 \hat{r}_1) x_{S^*,updated}^{t+h} \tag{8}$$

As suggested by prior works (Li et al. 2016, 2017), Stage 2 considers the pattern of residual errors regarding the water level. $x_{W^*}^{t+h}$ is used as the independent variable and $r_2$ is the dependent variable. Locally weighted scatterplot smoothing (LOWESS) is selected for the modeling because of its strong curve fitting ability, local linearity, and simplicity. The estimation of $\hat{r}_2$ is thus equal to $LOWESS(x_{W^*}^{t+h})$. Following the same idea in Stage 1, $c_2$ is calculated as the ratio of $MAPE_1$ and $MAPE_1 + MAPE_{reported}$, using the latest streamflow forecasting $x_{S^*,1}^{t+h}$. The updated forecasting and the residual error for the next stage are

$$x_{S^*,2}^{t+h} = (1 - c_2 \hat{r}_2) \, x_{S^*,1}^{t+h} \tag{7}$$

$$r_3 = x_{S^*,measured}^{t+h} - x_{S^*,2}^{t+h} \tag{8}$$

Stage 1 and 2 are designed to model the pattern in an explicit way. The advantage of this practice is that it is based on relevant theory or insight for the data and is easy to interpret and visualize, while the disadvantage is the lack of the ability to model complex patterns such as nonlinearities. Therefore, Stage 3 uses XGBoost for residual error learning, which is often used in machine learning to model complex nonlinear patterns. Forecasted water level $x_{W^*}^{t+h}$ and the water level delta $\Delta x_{W^*}^{t+h}$ are used as features for the XGBoost, and $r_3$ is the target of forecasting. When implementing XGBoost, as the number of data points for the model training in this stage is usually below 100, due to the limited number of streamflow field measurements, the bootstrapping method is utilized to stabilize the forecasting by XGBoost. The final streamflow forecasting by the proposed method of this study is:

$$x_{S^*,3}^{t+h} = x_{S^*,2}^{t+h} - XGB(x_{W^*}^{t+h}, \Delta x_{W^*}^{t+h}) \tag{9}$$

*3.4 Evaluation metrics and baseline models*

The performance of the proposed model is evaluated by metrics commonly used in streamflow forecasting and machine learning fields, including the MAE, MAPE, RMSE, bias, NSE, and correlation coefficient (CC). The model has higher accuracy when NSE and CC are higher, bias is closer to zero, and other metrics are lower. The formulations of these metrics are as below. In addition, to evaluate the model performances during floods, bias is extended to peak bias, peak percentage bias, and peak time bias. Peak bias is the bias during the flooding period when the water level is above the 'action' threshold defined by NOAA. Peak percentage bias is the peak bias divided by the water level, and peak time bias is the difference between the forecasted time of the highest streamflow during the flooding period and the actual peak time.

$$\text{MAE} = \frac{1}{n}\sum_{i=1}^{n}|y_i - \hat{y}_i| \tag{10}$$

$$\text{MAPE} = \frac{100\%}{n}\sum_{i=1}^{n}\left|\frac{y_i - \hat{y}_i}{y_i}\right| \tag{11}$$

$$\text{RMSE} = \sqrt{\frac{1}{n}\sum_{i=1}^{n}(y_i - \hat{y}_i)^2} \tag{12}$$

$$\text{bias} = \frac{1}{n}\sum_{i=1}^{n}(\hat{y}_i - y_i) \tag{13}$$

$$\text{NSE} = 1 - \frac{\sum_{i=1}^{n}(y_i - \hat{y}_i)^2}{\sum_{i=1}^{n}(y_i - \bar{y})^2} \tag{14}$$

$$\text{CC} = \frac{\sum_{i=1}^{n}(\hat{y}_i - \overline{\hat{y}_i})(y_i - \bar{y})}{\sqrt{\sum_{i=1}^{n}(\hat{y}_i - \overline{\hat{y}_i})^2 \sum_{i=1}^{n}(y_i - \bar{y})^2}} \tag{15}$$

, where $n$ is the number of data points in the evaluation set, $y_i$ is the $i^{th}$ ground truth value, $y_i$ is the forecasted value, $\bar{y}$ is the mean of ground truth values, and $\overline{\hat{y}_i}$ is the mean of forecasted values.

To validate that the graph convolutional RNN base model is effective, some baseline models are used. The Persistence model provides naïve forecast of streamflow. It simply uses the latest reported streamflow to forecast the streamflow $h$ time points later in this study. Linear regression, MLP, and XGBoost were also adopted, as they are among the most commonly used shallow machine learning models in prior studies and the machine learning field. The structure of these machine learning models is not designed specifically to capture the spatial or temporal patterns in the data. As a more advanced machine learning model, GRU was designed for the time series data and temporal patterns. It was also selected as a baseline model. Note the base model introduced in the Method section forecasts water level, which is further converted to streamflow by the corresponding rating curve. The reason for employing this modeling practice is that the streamflow, which is converted from the rating curve power function, has a larger range than the water level. The potential benefit of using a graph convolutional RNN plus rating curve (DCRNN+RC) approach is that it reduces the range of data points, which is favorable for the gradient descent optimization. Therefore, the DCRNN that directly forecasts streamflow is also used as a baseline model. All the models are fed with the same features and train-validation-test splits.

**4 Results**

*4.1 Rating curve errors amplified during flooding*

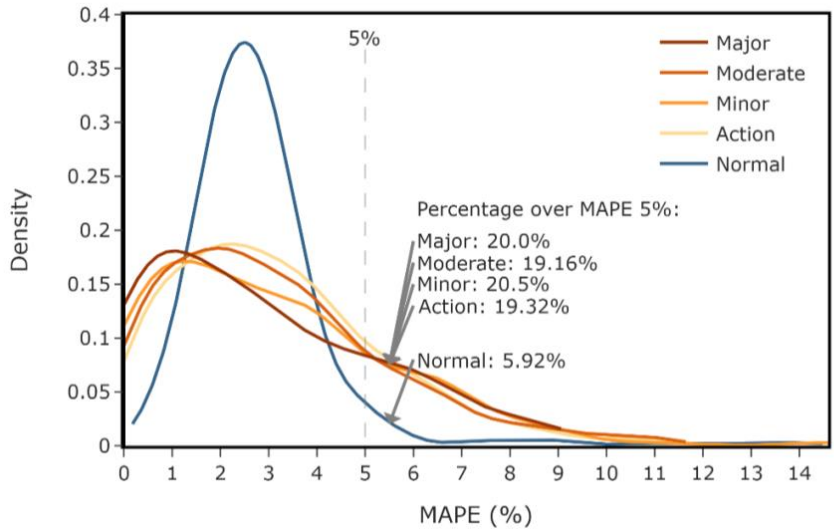

**Fig. 2.** Probability density functions of reported streamflow's mean absolute percentage error regarding measured streamflow for normal water levels, and action, minor, moderate, and major flood stages. The probability density functions were estimated using the Kernel Density Estimation method. The annotations show the probability that Mean Absolute Percentage Error (MAPE) exceeds 5%.

Before implementing the proposed machine learning method, we calculated the differences between the reported streamflow and measured streamflow for the 1,548 selected United States Geological Survey (USGS) gauging stations across the United States, using the data collected from the USGS National Water Information System. The Mean Absolute Percentage Error (MAPE) of the reported streamflow regarding the measured streamflow was calculated for each gauging station. The data points for the calculation are filtered with a varying water level threshold, indicating normal water level and four flood stages, which produces multiple groups of MAPE metrics. **Fig. 2** illustrates the probability density function of MAPE for these groups. Although there is no significant difference in the MAPE distributions of the "action", "minor", "medium", and "major" flood stages, these distributions are obviously different from that of "normal" water levels. The probability of high MAPE increases dramatically from "normal" to flood stages. Specifically, if 5% is used as a threshold, the probability that the "normal" MAPE exceeds 5% is only less than 6%, but the probability of the MAPE of the other distributions exceeding the 5% threshold is about 20%. This suggests that about 20% of the gauging stations with flood events are providing streamflow that differs from the measured values by 5% under flood conditions. Considering that the MAPE of short-term streamflow forecasting in prior studies is usually below 10% (Lin et al. 2021; Nguyen et al. 2022), the 5% bias in the dataset could potentially increase the error by 50% and is therefore not negligible. It supports the basic objective of the present study, which is to reduce the negative influence of the average 5% error on the forecasting accuracy.

*4.2 Graph convolutional recurrent neural network-based streamflow forecasting*

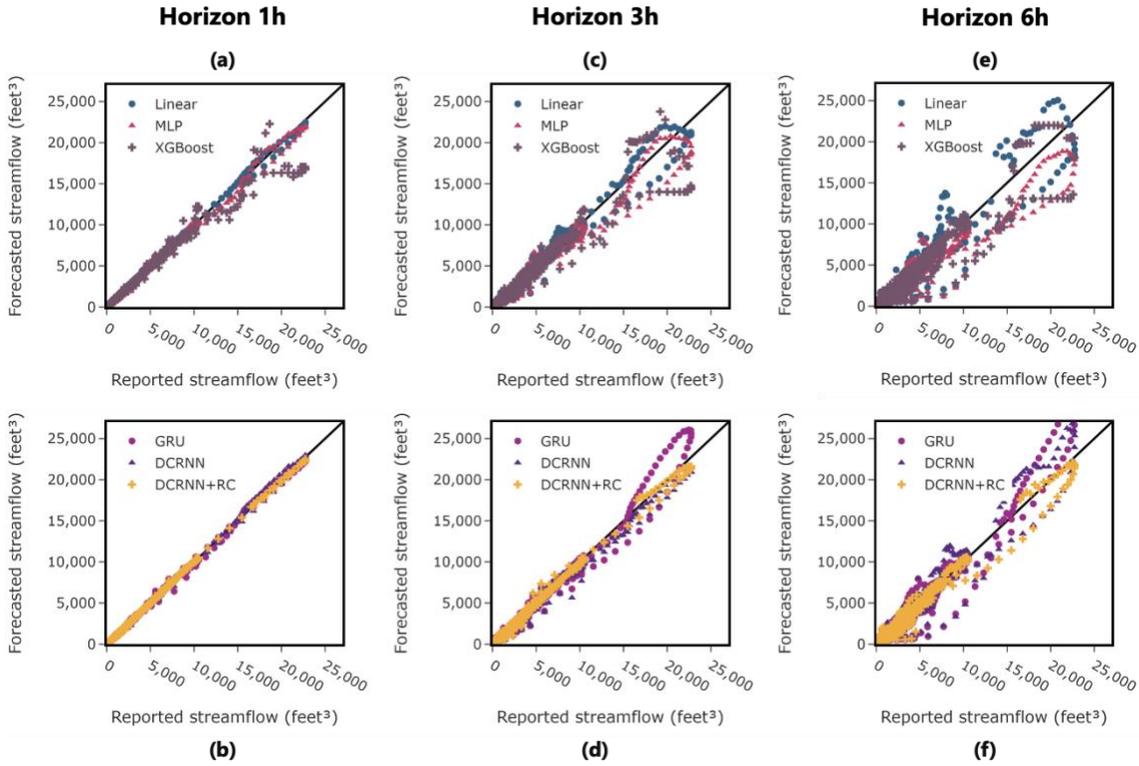

**Fig. S1.** The forecasted streamflow $x_{S^*}^{t+h}$ and the reported streamflow. (a)-(b) are the performances of the forecasting with a 1h horizon, (c)-(d) are the performances of the forecasting with a 3h horizon, and (e)-(f) are the performances of the forecasting with a 6h horizon. (a), (c), and (e) are the performances of shallow machine learning models (Linear, MLP, and XGBoost), and (b), (d), and (f) are from the performances of deep learning models (GRU, DCRNN, and DCRNN+RC). The result is based on the models for gauging station USGS 01573560.

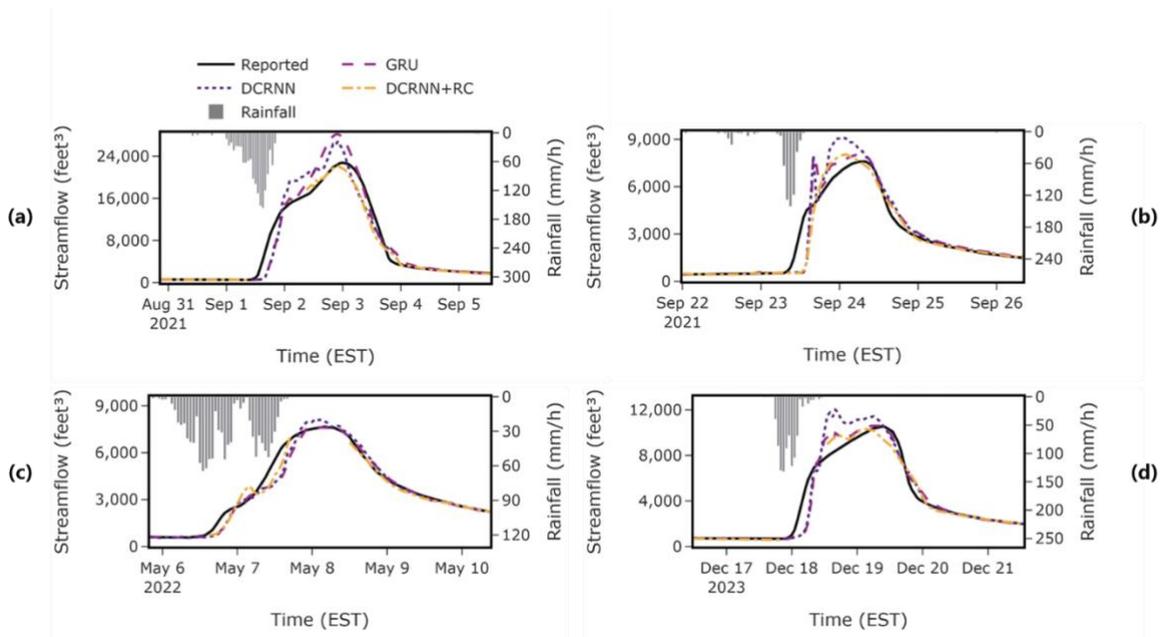

**Fig. 3.** The reported streamflow and the forecasted streamflow $x_{S^*}^{t+h}$ by the deep learning models during flood events. The forecasting horizon for all sub-figures is 6h. The result is based on the models for the gauging station USGS 01573560.

To solve the above-mentioned issue, the method presented in **Fig. 1** was implemented. **Fig. S1** illustrates the forecasting accuracy of the base model as well as the baseline models concerning the reported streamflow. The result shows that the accuracy of Gated Recurrent Unit (GRU) is obviously better than commonly used machine learning baselines (i.e., linear regression, multilayer perceptron, and XGBoost). However, GRU is outperformed by diffusion convolution recurrent neural network (DCRNN) and its variant DCRNN+ rating curve (DCRNN+RC), which outputs forecasted water level first and then converts it to forecasted streamflow using rating curves. **Table S1** compares the performances of the models using metrics. DCRNN and DCRNN+RC have the largest number of best metrics (25 out of 30 in total). Because the motivation for this study is the short-term streamflow prediction for flood forecasting, the performance of the base model, as well as the deep learning baseline models, during flood events were further evaluated. The periods of flood events were determined by the periods when the water level exceeds the "action" stage defined by the National Oceanic and Atmospheric Administration. The "action" stage is when the water is close to or slightly above the riverbank, which means that immediate flood management action is necessary. As shown in **Fig. 3**, there are four flood events at the gauging station USGS 01573560 during the test period. All three deep learning models evaluated can well forecast the general trend of the streamflow 6 hours in advance. However, DCRNN+RC's curve is closer to the black reported streamflow line in most cases (e.g., the peak time in **Fig. 3(a), (b)**, and the rising period of **Fig. 3(c)**). The advantage of DCRNN+RC is clearer in **Fig. 3(b)** and **(d)**, where the baseline models' curves have false spikes in the rising period (i.e., Sep. 23rd afternoon in **Fig. 3(b)** and Dec. 18th in **Fig. 3(d)**), but DCRNN+RC does not. **Table S2** demonstrates the corresponding metrics. DCRNN+RC has a clear advantage in all metrics except for bias. Overall, the base model can forecast the reported streamflow with higher accuracy than those of commonly used models within a horizon of several hours. Although the base model could forecast reported streamflow with higher accuracy, the performance of forecasting measured streamflow has not been evaluated yet. **Fig. S2**, **Table S3**, and **Table S4** compare the metrics of the base model regarding the reported streamflow and the measured streamflow. The performance with respect to the measured streamflow will decrease. This is expected because the training target of the base model is the reported streamflow, not the measured streamflow.

*4.3 Residual error learning and reduction*

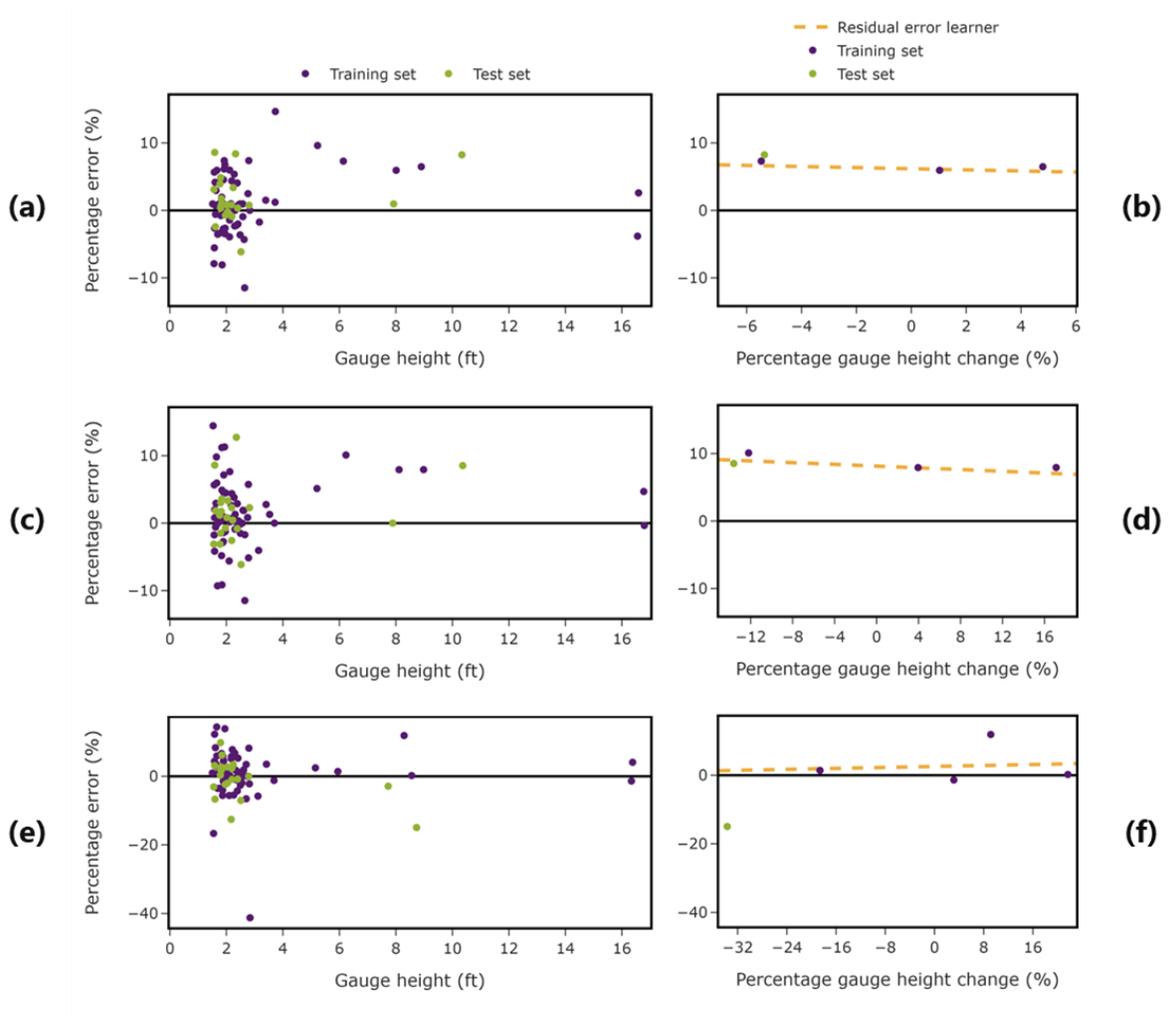

**Fig. 4.** Percentage residual error $r_1$ versus the water level $x_{W^*}^{t+h}$ and the delta of water level $\Delta x_{W^*}^{t+h}$ and the Stage 1 linear regression model with varying forecasting horizons. (a)-(b) are the results for the 1-hour horizon, (c)-(d) are the results for the 3-hour horizon, and (e)-(f) are the results for the 6-hour horizon. (a), (c), and (d) are the percentage error versus the water level. (b), (d), and (f) are the percentage error versus the water level delta and the Stage 1 regression model. The result is based on the models for gauging station USGS 01573560.

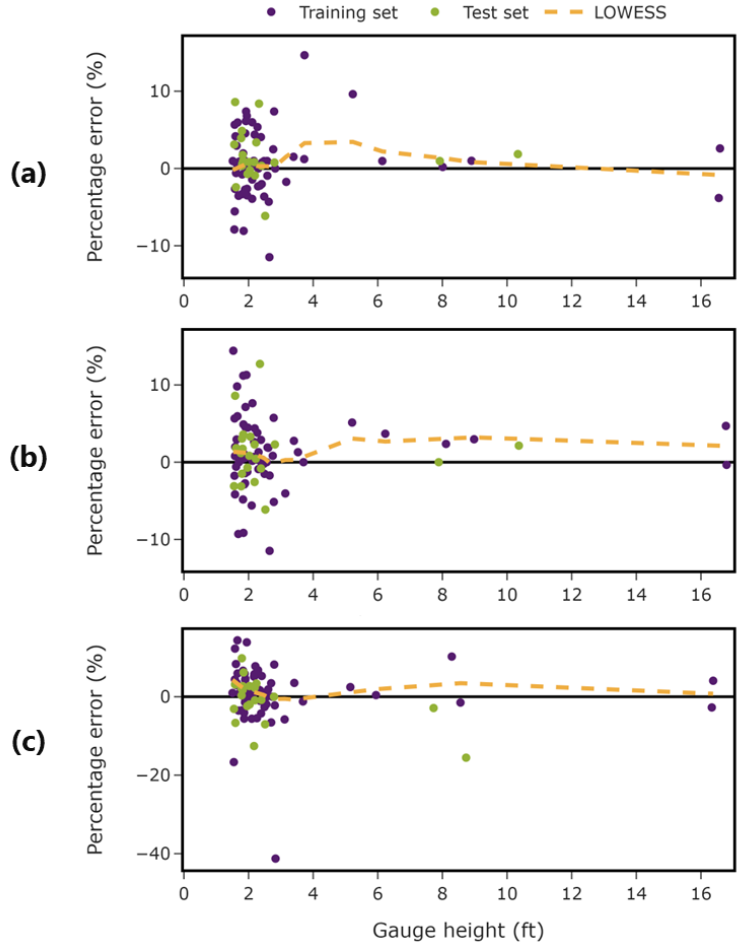

**Fig. 5.** Percentage residual error $r_2$ versus the water level $x_{W^*}^{t+h}$ and the Stage 2 LOWESS model with varying forecasting horizons. (a) is the result for the 1-hour horizon, (b) is the result for the 3-hour horizon, and (c) is the result for the 6-hour horizon. The result is based on the models for gauging station USGS 01573560.

The results from **Fig. S2**, **Table S3**, and **Table S4** indicate that 1) the reported streamflow error from rating curves is not negligible, and 2) the importance of the residual error due to the rating curves declines as the forecast horizon increases. The **Methods** section presents a method to perform residual error learning and reduction considering these two issues. As a result, **Fig. 4** shows the residual error pattern to be learned by the Stage 1 computation versus water level and water level change for USGS 01573560. The "action" stage for the gauging station used in **Fig. 4** is 6 feet. As seen in **Fig. 4(a)**, although the residual error pattern is not obvious when the water level is below the "action" stage, there is a clear decreasing trend after the water level approaches it. The overall residual error pattern is quadratic-like. **Figure 4(b)** shows the data points filtered by $e_{action}$. There is a linear-like decreasing trend as the water level change rises. Note that the patterns change as the forecasting horizon increases. Although the above-mentioned patterns still hold for the 3-hour horizon, the residual error is close to being random at the 6-hour horizon. This result suggests that the pattern of residual errors is decaying, and it echoes the findings from **Fig. S2**, **Tables S3,** and **Table S4**. This result supports the research practice of using $c_1$ and $c_2$ to adjust the degree of applying residual error reductions in the Methods. The findings from **Fig. 4** still hold in the analysis for USGS 0332000 (shown in **Fig. S3**).

**Fig. 5** demonstrates the residual errors after applying the linear models in **Fig. 4(b), (d)**, and **(f)**. While the residual error is not obviously reduced for the 6-hour horizon case, it is significantly reduced for the 1-hour and 3-hour horizons, when the water level is above the "action" stage. The LOWESS model is then applied to the reduced residual errors. The trained LOWESS models are visualized in **Fig. 5**. Although the pattern of residual errors at low water levels is not obvious, there is a slight downward trend as seen in the LOWESS models in **Fig. 5(b)** and **(c)**. At high water levels, the LOWESS models are close to the training and test data points in **Fig. 5(a)** and **(b)**. **Fig. S4** illustrates a similar analysis for USGS 03320000.

The performance metrics in **Tables S5-S8** demonstrate the improved model accuracy more clearly. Taking the overall metrics for USGS 01573560 (shown in **Table S5**) as an example, in the case of a 1–3-hour forecasting horizon, the metrics were improved from DCRNN+RC to DCRNN+RC+Stage 1, 2, except for the MAPE of the 3h horizon. After the 3h horizon, the accuracy is improved after Stage 1 but decreases slightly in subsequent stages. But in the vast majority of cases, the forecasting processed by the residual error reduction is more accurate than the original forecasting by the DCRNN+RC. XGBoost was then used in Stage 3 to capture the possible nonlinear patterns. As shown in **Tables S5** and **S6**, the XGBoost model can further improve the accuracy at the 1-hour horizon, especially the accuracy of the forecasting during floods. It can improve part of the metrics when the horizon is 3 hours but will slightly reduce the accuracy at other horizons. In the case of USGS 03320000, despite very few exceptions, the overall mean absolute error (MAE), MAPE, root mean squared error (RMSE), and Nash-Sutcliffe efficiency (NSE) were improved across all forecasting horizons from DCRNN+RC to DCRNN+RC+Stage 1, 2, 3 (shown in **Table S7-S8**). In particular, MAPE decreases after each calculation from Stage 1 to 3.

## 5 Discussion and Conclusions

This study evaluates the difference between reported and field-measured streamflow provided by the gauging stations across the U.S. The number of gauging stations selected was approximately 15% of the active gauges operated by the USGS. Considering that the USGS provides the largest and most widely used streamflow dataset in the U.S. and that previous studies (Coxon et al. 2015) used hundreds of gauging stations for similar purposes, the gauging stations selected are comprehensive. The corresponding result illustrates that the errors introduced by the rating curves are less during non-flood periods. However, about 20% of the gauging stations are providing streamflow data with more than 5% MAPE regarding the measured streamflow during flood periods. Considering the MAPE of streamflow forecasting in prior studies is usually below 10% (Lin et al. 2021; Nguyen et al. 2022), 5% MAPE in the training data itself is of significant concern. The results of this evaluation highlight the inaccuracy of rating curves and their potential impact on the streamflow dataset. Because the dataset under study is the largest streamflow dataset in the U.S. and underpins many of the machine learning models for streamflow forecasting, the finding suggests that researchers and practitioners reevaluate the effectiveness of streamflow forecasting models and the flood management decisions built on that.

Additionally, this study builds a more accurate neural network model to perform short-term streamflow forecasting. Many prior studies with the use case where multiple upstream gauges exist overlooked the importance of spatial pattern recognition(Nguyen et al. 2022). Although the Long Short-term Memory model has been widely used in prior research (Cui et al. 2021; Haznedar et al. 2023), the recurrent neural network architecture was not designed to capture spatial patterns. The graph convolution operation used in this study can better process the water level and the streamflow information from multiple gauges and improve the accuracy of the streamflow prediction at the target location. Built on the neural network, this study proposes a method for learning and reducing the residual error induced by rating curves. It follows previous studies that used water levels to model the residual error (Li et al. 2016) and also make use of the magnitude of water level change, which is rarely used in related studies. The improvements in the forecasting accuracy after the Stage 1 calculation echo the conclusion from a prior study(Wolfs and Willems 2014) that the water level change would lead to the residual error through rating curves. Our study is helpful for enhancing the accuracy of short-term streamflow forecasting during floods. It is expected to help mitigate overestimating or underestimating streamflow during flood periods and more accurately inform flood management measures, such as controlling the water release of upstream dams.

To conclude, the authors argue that the differences between the reported and the measured streamflow caused by imperfect rating curves can negatively affect the accuracy of short-term streamflow forecasting. The method proposed in this study can forecast the reported streamflow with advanced accuracy. It can also learn and reduce the residual error caused by the rating curve by utilizing a limited number of measured streamflow. Among the three stages of computations proposed, Stage 1 is the most effective, and Stages 2 and 3 can further reduce the residual error in the part of the forecasting horizons. Overall, the proposed method is effective at all prediction horizons from 1-6 hours, with better accuracy at the 1-3 hour horizons. The present study provides a more accurate method for short-term forecasting of flood streamflow, which is a fundamental technique to provide reference information for river flood response decision-making, such as flood water storage and release at flood peaks.